\documentclass{article} 
\usepackage{iclr2019_conference,times}

\usepackage{amsmath,amsfonts,bm}

\def\eqref#1{equation~\ref{#1}}

\def\1{\bm{1}}

\DeclareMathAlphabet{\mathsfit}{\encodingdefault}{\sfdefault}{m}{sl}
\SetMathAlphabet{\mathsfit}{bold}{\encodingdefault}{\sfdefault}{bx}{n}

\usepackage{booktabs}
\usepackage{hyperref}
\usepackage{url}

\usepackage{amsmath}
\usepackage{graphicx}
\usepackage{multirow}
\usepackage{subfig}

\newcommand{\task}{\mathit{tsm}}
\newcommand{\obs}{\mathit{obs}}
\newcommand{\env}{\mathit{sem}}
\newcommand{\oembed}{\mathit{E}^{\obs}}
\newcommand{\tembed}{\mathit{E}^{\mathit{task}}}
\newcommand{\envrnn}{\mathit{RNN}^{\env}}
\newcommand{\trnn}{\mathit{RNN}^{\task}}
\newcommand{\fpol}{\mathit{F}^{\mathit{pol}}}
\newcommand{\fval}{\mathit{F}^{\mathit{val}}}
\newcommand{\fcomp}{\mathit{F}^{\mathit{comp}}}

\title{Continual and Multi-task Reinforcement Learning With Shared Episodic Memory}

\author{
  Artyom Y. Sorokin \\
  Moscow Institute of Physics and Technology\\
  Dolgoprudny, Russia \\
  \texttt{griver29@gmail.com} \\
  \And
  Mikhail S. Burtsev \\
  Moscow Institute of Physics and Technology\\
  Dolgoprudny, Russia \\
  \texttt{burcev.ms@mipt.ru} \\
}

\iclrfinalcopy

\begin{document}

\maketitle

\begin{abstract}
Episodic memory plays an important role in the behavior of animals and humans. It allows the accumulation of information about current state of the environment in a task-agnostic way. This episodic representation can be later accessed by down-stream tasks in order to make their execution more efficient. In this work, we introduce the neural architecture with shared episodic memory (SEM) for learning and the sequential execution of multiple tasks. We explicitly split the encoding of episodic memory and task-specific memory into separate recurrent sub-networks. An agent augmented with SEM was able to effectively reuse episodic knowledge collected during other tasks to improve its policy on a current task in the Taxi problem. Repeated use of episodic representation in continual learning experiments facilitated acquisition of novel skills in the same environment.
\end{abstract}


\section{Introduction}\label{sec:intro}

Humans and other animals use episodic memory to adapt quickly in complex environments \citep{kumaran2016what_agents_need, lake2017building, hippocampus_learn_memory_1995}. A striking feature of animal behaviour is ability to achieve several different goals in the same environment. Both of these features of adaptive behavior are being actively studied in the field of reinforcement learning \citep{p_distilation_rusu2015, gated_grounding_doom_chaplot2017,distral_teh2017, memory_mincraft_oh2016, neural_map_parisotto2017, relational_rnn_santoro2018, progress_compress_schwarz2018}. However, the majority of the studies consider them in isolation, focusing either on episodic memory \citep{model_ep_control_blundell2016,neural_ep_control_pritzel2017, relational_rnn_santoro2018} or on learning several policies for achieving different goals \citep{meta_shared_hrl_frans2017,dosovitskiy2016pred_future, lifelong_minecraft_tessler2017, progress_compress_schwarz2018}. Yet, the content of the episodic memory can be useful not only for a single task, but for completing multiple consecutive tasks, in addition to the general acquisition of new skills. For example, one can imagine a robotic home assistant instructed to retrieve a certain object. If the robot has encountered this object during a past house cleaning and recalls it, then this memory can greatly facilitate locating the requested object. 

In this work, we propose a deep neural architecture able to store the episodic representation of an environment to improve the solution of multi-task problems and facilitate continual learning of new skills.

\section{Related Work}\label{sec:related_works}


One of the most popular general approaches to multi-task learning is to train an agent on all tasks simultaneously. This  method is called batch multi-task learning \citep{lifelong_ml_chen2016}. In the field of deep reinforcement learning, it is often coupled with the weight sharing technique. In that case sub-task networks share part of their layers and the representation of the agent's current task is often fed in as additional input to the network \citep{stoch_nets_hrl_florensa2017,dosovitskiy2016pred_future, beating_atari_kaplan2017}. Weight sharing allows to generalize experience over tasks and facilitates the learning of individual tasks \citep{inro_rl_transfer_taylor2011}. A remarkable extension of this approach is the representation of sub-tasks with a descriptive system \citep{prog_agents_denil2017}. Previous work on neuroevolution in a multi-task stochastic environment \citep{lakhman2013neuroevolution} demonstrated that agents evolve representation for episodic memory and use it in behavior. Several studies have been done on the mapping of natural language task descriptions into sequences of actions \citep{gated_grounding_doom_chaplot2017, map_instructions_rl_misra2017}.  Another work in that area \citep{beating_atari_kaplan2017} used a sequence of instructions to guide the agent in Montezuma's Revenge game.
However, the majority of current research in the field is focused on the isolated execution of sub-tasks, ignoring the transfer of episodic memory between sub-tasks.




In the past couple of years, numerous works have been completed on adding memory to deep RL architectures \citep{a3c_mnih, drqn_hausknecht2015}. One direction of research is to improve the agent's ability to store relevant memory about the state of environment \citep{neural_map_parisotto2017, memory_mincraft_oh2016, relational_rnn_santoro2018}. Alternatively, memories about recent state transitions  can be used to facilitate rapid learning \citep{model_ep_control_blundell2016, neural_ep_control_pritzel2017}.

It has been demonstrated that recurrent neural networks (RNNs) are capable of meta-learning \citep{thrun1998learning2learn, oneshot_mann_santoro2016}. Meta-learning in this case typically refers to the interaction of two learning processes. Slow adaptation when the weights of the neural network gradually learn persistent regularities in the environment. And fast dynamics of the recurrent network to adapt for rapid changes in the environment. Recently, this approach was extended to the RL setting \cite{meta_rl_wang2016, meta_rl_duan2016}. In another work \citep{sim2real_peng2018}, a similar training technique helped to transfer a policy learned in simulation to a physical robot. While current focus of meta-RL with recurrent architectures is mainly on the adaptation of one policy to different variations of the environment we will consider a joint adaptation of several goal-oriented policies via shared memory.

\section{Shared Episodic Memory for Multi-Task Reinforcement Learning}

In this work, we use two ideas to facilitate the transfer of useful episodic representation between multiple sub-task policies. The first is an introduction of two separate recurrent sub-networks (1) for the environment and (2) for task-specific memories. The second is to use meta-learning setting \citep{meta_rl_duan2016, meta_rl_wang2016, meta_shared_hrl_frans2017} to optimize the agent over a series of tasks in the same environment. 



\label{sec:meta_rl_procedure}
Traditionally in multi-task reinforcement learning setting, a new task is selected at the beginning of each episode so that one episode corresponds to one task. An agent then simultaneously interacts with several instances of the environment and updates policy using samples collected for different tasks. This procedure is ineffective in storing a representation for more than one task. To make episodic memory useful we train a multi-task agent in a setting similar to the one used in meta-RL \citep{meta_rl_wang2016, meta_rl_duan2016}.

In our study, training consisted of episodes lasting $T$ steps. For every episode, the environment was modified to some extent, i.e. locations of walls, targets, or objects. At the beginning of an episode, a task was randomly selected.  If the task was completed by the agent, a new task was activated, but the state of the environment remained the same. Upon task completion, the agent received a reward and a "completion" signal shared among all tasks. The agent optimized the cumulative reward for all $T$ steps.

Thus, the more tasks an agent completed in the available time budge more reward it received. This training mode encourages the agent's neural network not only to learn suitable policies for tasks but also to share between them a memory about the state of the environment.

\begin{figure}[h]
\centering\includegraphics[width=0.9\linewidth]{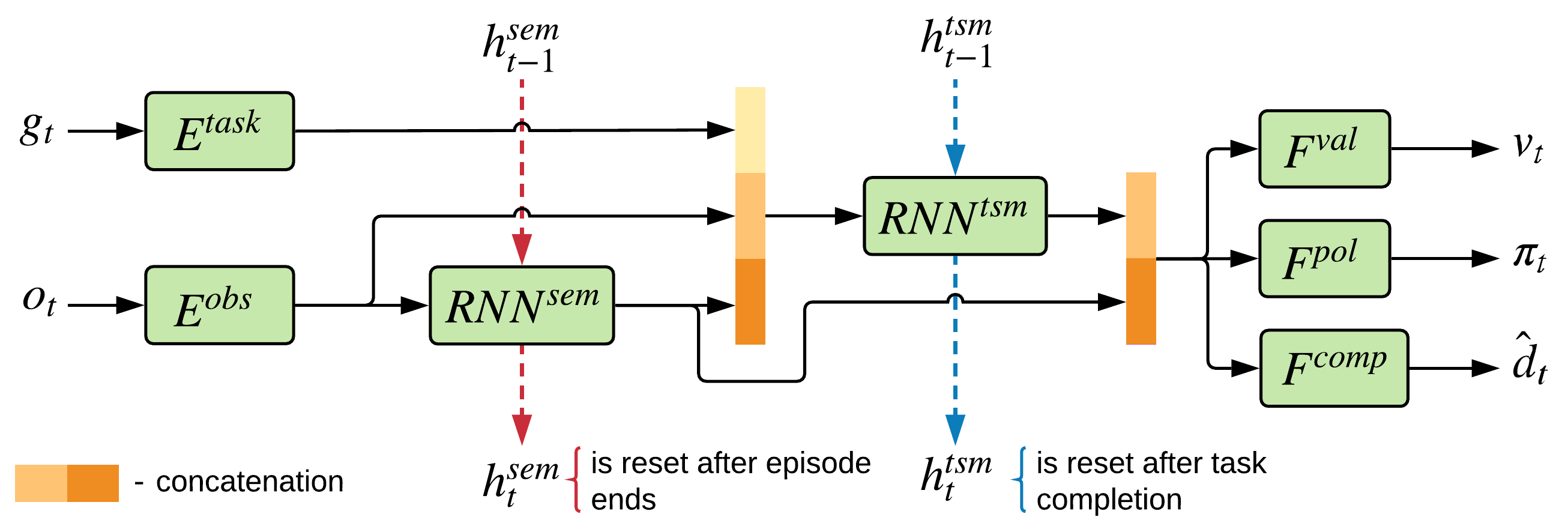}

\caption{A diagram of the SEM-A2C network architecture. The green blocks are trainable modules. Dashed blue and red lines indicate the flow of information through the hidden states of the recurrent memory modules over time. SEM-A2C encodes current observation $o_t$ and adds it to episodic memory $\envrnn$. The task specific memory $\trnn$ is updated with the embeddings of task ids $g_t$ and $o_t$  as well as the content of episodic memory $h_{t}^{\env}$. Finally, episodic and task specific memories are fed to policy $F^{pol}$, value $F^{val}$ and task completion $F^{comp}$ heads.}
\label{fig:paac_mt_network}
\end{figure}

To train the agent, we used the Parallel Advantage Actor Critic (\emph{A2C}) algorithm \citep{a3c_mnih, paac_clemente2017}. Our proposed network architecture with \emph{shared episodic memory} (\emph{SEM-A2C}) is presented in Figure \ref{fig:paac_mt_network}.  Instead of LSTM layer of the recurrent A2C, we introduce separate memory sub-networks for the environment state and task. At each step $t$, the network receives the current observation $o_t$, and the task identifier $g_t$. Observation $o_t$ is processed by the observation encoder $\oembed$, and identifier $g_t$ by task embedder $\tembed$. In our experiments, $\oembed$ is a two-layered convolutional network, and $\tembed$ is an embedding matrix that stores trainable task embeddings in its rows.

The core of the proposed architecture consists of  $\envrnn$ and $\trnn$ recurrent sub-networks. $\envrnn$ takes observation embedding $\oembed(o_t)$ and returns its hidden state $h_{t}^{\env}$, which is reset to zero values only at the end of each episode. $\trnn$ takes the same input as $\envrnn$ as well as $h^{\env}_t$ and task embedding $v_{g_t}$. Unlike $\envrnn$ the hidden state $h^{\task}_t$ of the $\trnn$  is reset after the completion of the current task. 
The idea is that $\envrnn$ is responsible for capturing and storing a task-agnostic representation of the environment state, and $\trnn$ encodes a task specific representation. In contrast, $\envrnn$ has no knowledge of the current task, but is continuously updated over a longer period, which would correspond to several tasks in the batch multi-task setting. 

In our experiments, we use a single LSTM layer \citep{lstm_hochreiter1997}  for $\envrnn$.  For $\trnn$, we use a factorized LSTM layer (F-LSTM) to generate weights of $\trnn$ on the fly with the task embedding vector $v_{g_t}$ and the multiplication of three smaller matrices (see suppl. \ref{sec:f_lstm}).

Outputs of $\trnn$ and $\envrnn$ are concatenated and fed to three separate heads implemented as fully-connected layers.  The first two are standard actor-critic heads where $\fval$ predicts the state value function and $\fpol$ generates the probabilities of the actions. The last head $\fcomp$ predicts the probability of task completion (see suppl. \ref{sec:sem_paac_formula}).

Parameters are updated in SEM-A2C in the same way as in A2C \citep{paac_clemente2017} and A3C \citep{a3c_mnih} algorithms. To learn the task completion prediction $\hat{d}_t$ we use cross-entropy loss.

\section{Experiments}

We studied SEM-A2C performance in randomized grid-worlds implemented on top of the Mazebase engine \citep{mazebase_sukhbaatar2015}. For multi-task experiments, (sect. \ref{sec: multi_task_experiments}) an agent was trained for the sub-tasks of the Taxi problem \citep{max_q_dietterich2000}: reach passenger (\textit{Reach}(P)), pick up passenger (\textit{Pickup}(P)), reach destination (\textit{Reach}(D)), and drop off passenger (\textit{Dropoff}(P)). For continual learning experiments (sect. \ref{sec:continual_experiments}), we added a new cargo object on the map and three associated sub-tasks.

The map was filled with randomly placed walls and ponds. To make the problem harder an agent can only see objects in a small 7x7 cell grid which surrounds it, and has no direct information about the coordinates of the passenger, cargo and destination (see Figure \ref{fig:heatmaps}).

In our study, we consider sub-tasks as separate goals for the agent. If the agent completes its current sub-task, the next sub-task is selected from those which may follow logically in the current state. We place the agent and every pickable object at a new location on the map after completion of every two or three sub-tasks. However, the map and location of the target remain unchanged throughout all $T$ steps of the episode. Thus, it is beneficial for the agent to transfer acquired knowledge about the structure of the maze and location of the target between sub-tasks to facilitate their completion.

\subsection{Multi-task learning}

As a baseline we used a batch multi-task A2C (\emph{Multitask-A2C}) with weight sharing and task parametrization \citep{gated_grounding_doom_chaplot2017, prog_agents_denil2017, stoch_nets_hrl_florensa2017, dosovitskiy2016pred_future}. It received the same input as SEM-A2C including $g_t$, but instead of separate  $\trnn$ and $\envrnn$ sub-networks, it has a single LSTM layer with approximately the same number of parameters.  Multitask-A2C is trained via batch multi-task learning, where each episode corresponds to one task.

To evaluate the utility of episodic memory shared between the learned policies, we tested SEM-A2C and Multitask-A2C on the full taxi problem, where each agent had to perform the following sequence of tasks: \textit{Reach}(P), \textit{Pickup}(P), \textit{Reach}(D), \textit{Dropoff}(P). After dropping off a passenger at the target location a new one was spawned at a random location on the map. Each algorithm was previously trained for 80 million steps.

The results are presented in the Table \ref{tab:intertask_learning}. The table shows the number of steps required to complete tasks \textit{Reach}(P) and \textit{Reach}(D) depending on the order in which the tasks were performed in the episode. As can be seen from the table, SEM-A2C but not the Multitask-A2C baseline, manages to optimize the solution of \textit{Reach}(D) sub-task. After the agent with shared episodic memory discovers the target it can deliver future passengers to the same location $40\%$ faster. Due to episodic memory SEM-A2C solves the full Taxi problem with $20\%$ less steps than Multitask-A2C. 

\begin{figure}[h]
    \centering
    \subfloat[Multitask-A2C, \newline before visiting target \label{fig:heatmap_1}]{%
		\includegraphics[width=0.2\textwidth]{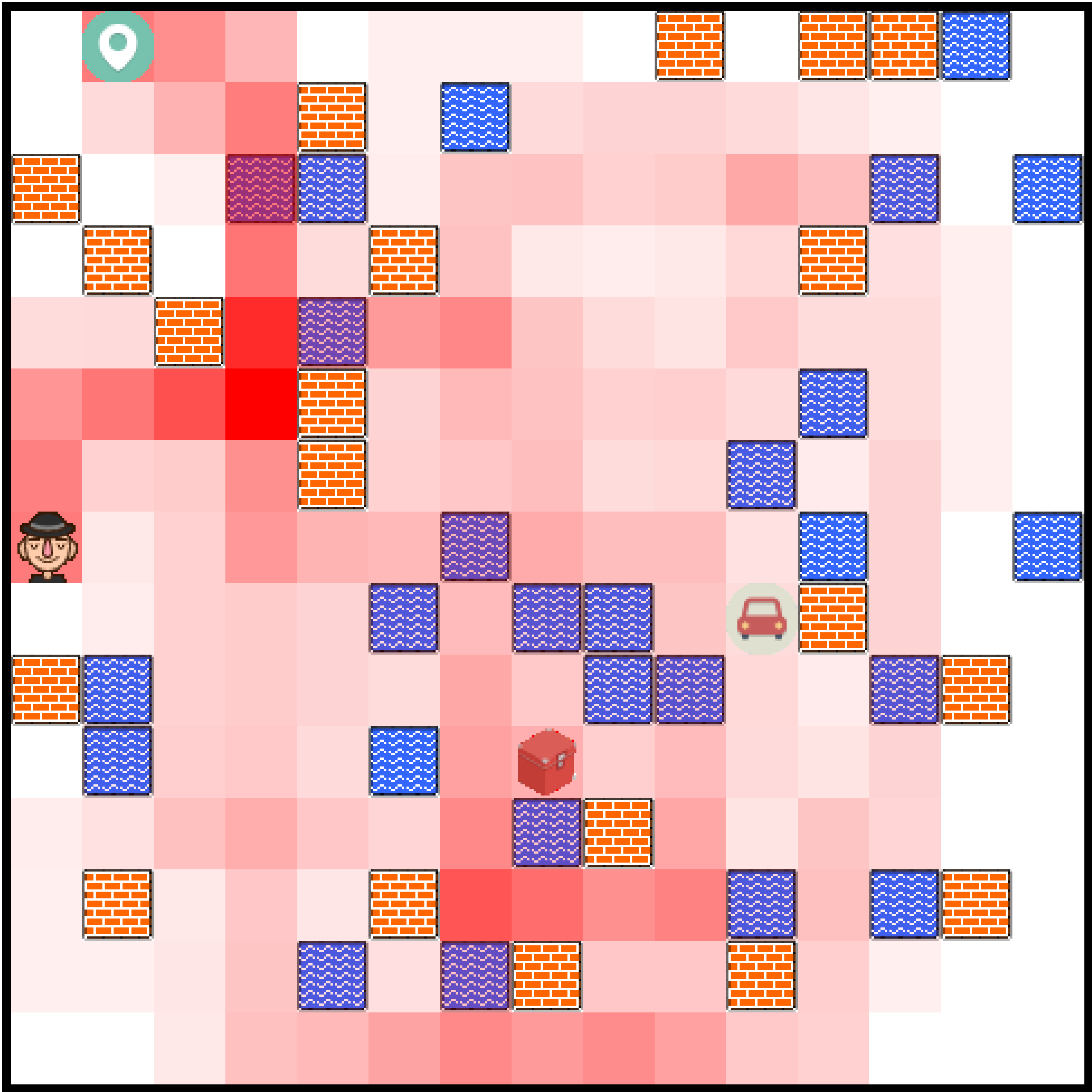}%
	}
    ~
    \subfloat[Multitask-A2C, \newline after visiting target \label{fig:heatmap_2}]{%
        \includegraphics[width=0.2\textwidth]{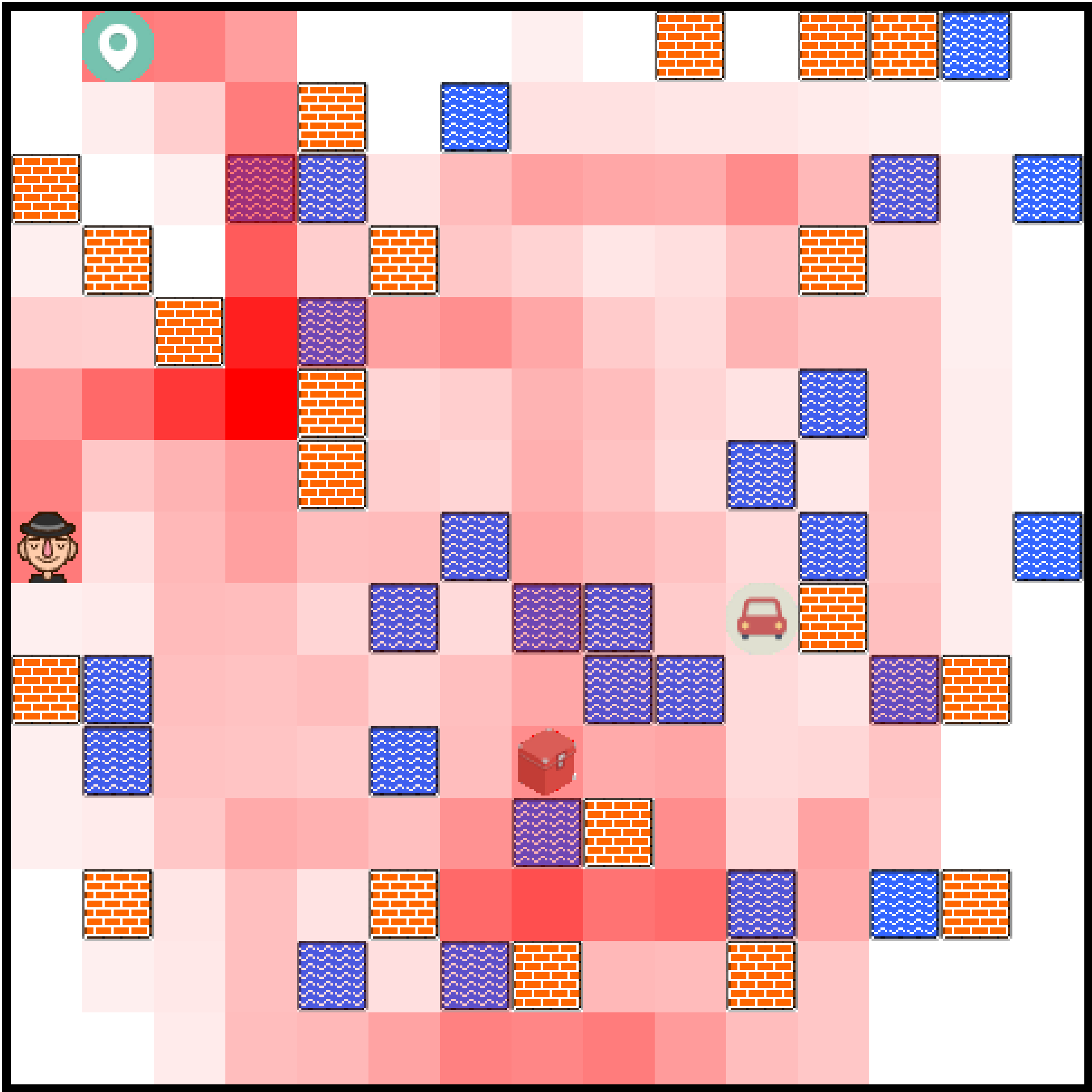}%
    }
     ~
     \subfloat[SEM-A2C, \newline before visiting target \label{fig:heatmap_3}]{%
        \includegraphics[width=0.2\textwidth]{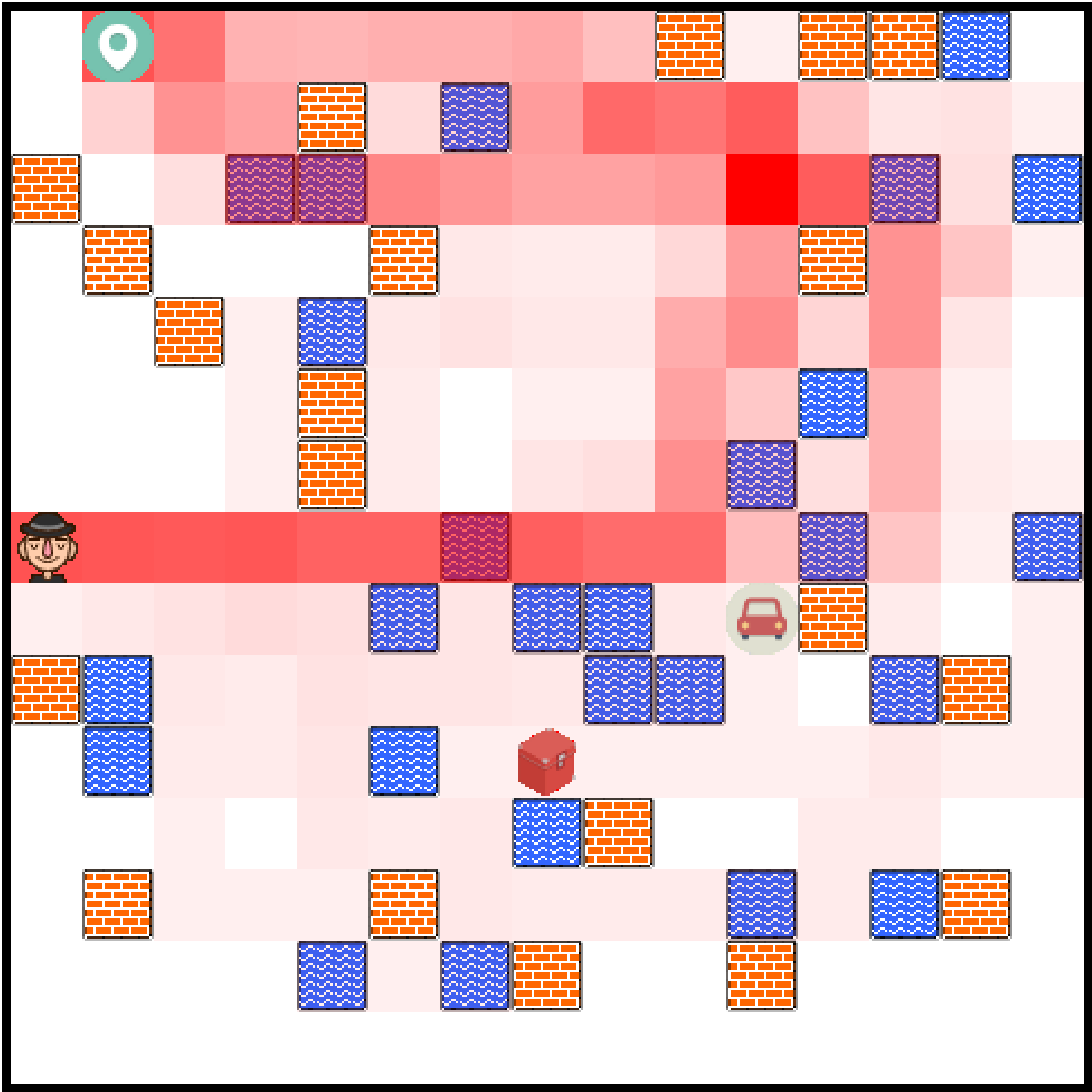}%
    }
    ~ 
    \subfloat[SEM-A2C, \newline after visiting target \label{fig:heatmap_4}]{%
        \includegraphics[width=0.2\textwidth]{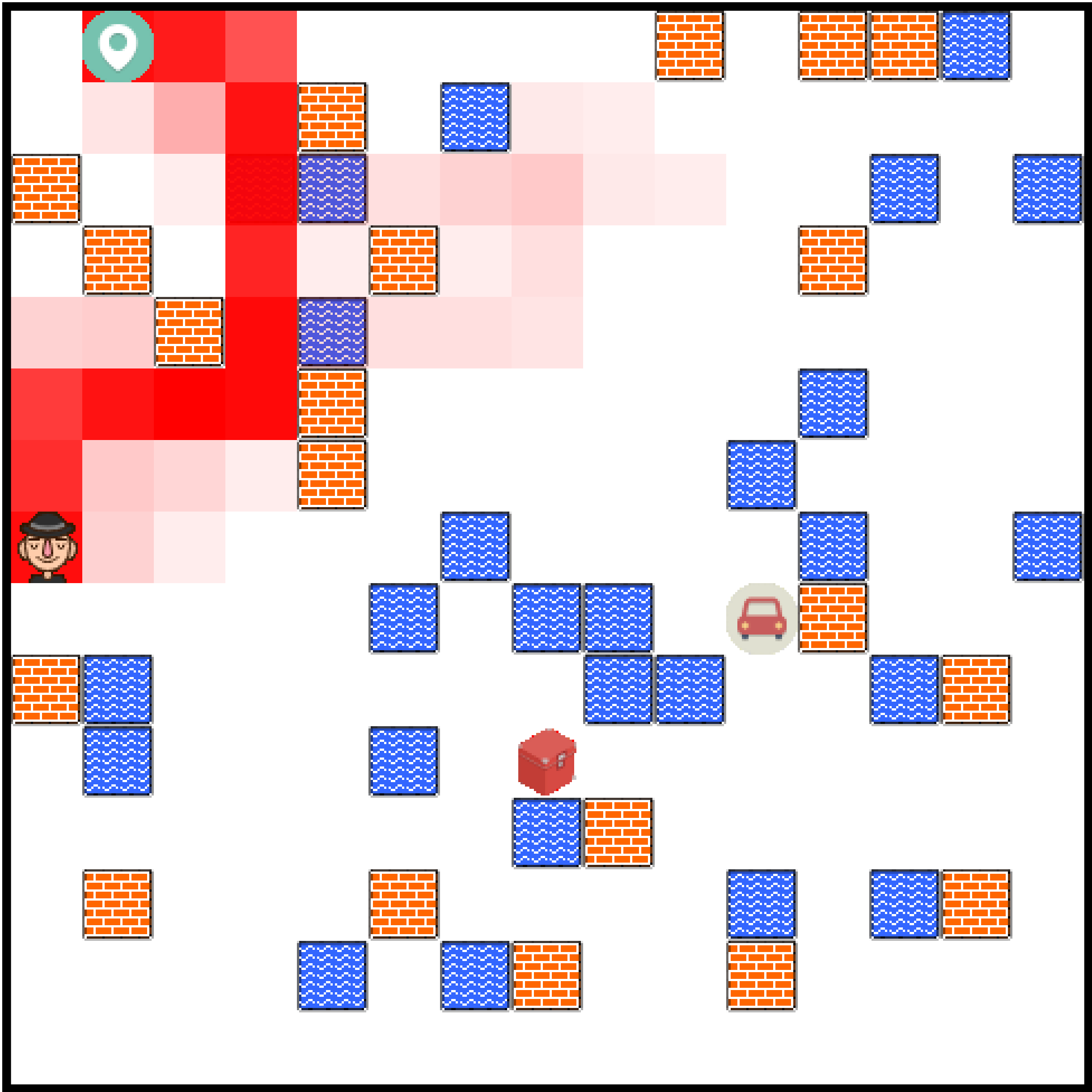}%
    }
	~
    \subfloat{%
        \raisebox{-0.13\height}{\includegraphics[width=0.105\textwidth]{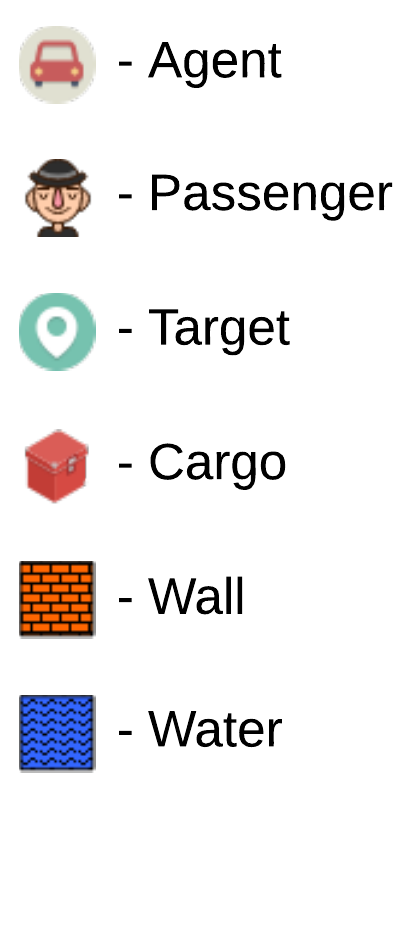}}%
    }
    \caption{The heatmap shows the relative frequency of the agent visiting each location during the sub-task of carrying the passenger to the target location(\textit{Reach}(D)) for Multitask-A2C and SEM-A2C. Each heatmap represents results averaged over 50 independent runs. The car icon shows location of the agent on the map at the beginning of the episode. Panels (a) and (c) show the behavior of agents that have never visited the target location before receiving the Reach(D) task. Panels (b) and (d) show the behavior of agents which visited the target location during the previous tasks in the episode.}\label{fig:heatmaps}
\end{figure}

 Figure \ref{fig:heatmaps} shows the difference between SEM-A2C and Multitask-A2C agents on a random fixed map. SEM-A2C drastically improves its policy by utilizing the experience obtained by performing previous tasks in this episode (Figures \ref{fig:heatmap_3}, \ref{fig:heatmap_4}). Conversely, the Multitask-A2C Baseline did not learn to account its experience from the previous tasks performed in the same environment (Figures \ref{fig:heatmap_1}, \ref{fig:heatmap_2}).

\begin{table}[ht]

  \caption{Episodic memory improves performance on the Taxi problem. Table shows the number of steps to complete sub-tasks over an episode on the 15x15 map. Values are averaged over 500 runs.}
  \label{tab:intertask_learning}
  \centering
  \begin{tabular}{ccccc}
    \toprule
    Appearance of sub-task & \multicolumn{2}{c}{Reach Passenger sub-task} & \multicolumn{2}{c}{Reach Target sub-task} \\
    \cmidrule(r){2-3} \cmidrule(r){4-5}
     in an episode & SEM-A2C & Multitask-A2C & SEM-A2C & Multitask-A2C \\
    \midrule
    $1^{st}$  & 22.67  & 23.61  & 24.26 & 25.38 \\
    $2^{nd}$  & 22.38  & 21.34  & 15.04 & 26.41 \\
    $3^d$  & 25.53  & 21.84  & 15.19 & 25.33 \\
    $4^{th}$ & 25.14  & 24.48  & 16.09 & 25.21 \\
    $5^{th}$ & 25.5   & 22.40  & 14.80 & 24.02 \\
    \bottomrule
  \end{tabular}
\end{table}

\label{sec: multi_task_experiments}

\subsection{Continual learning}

\label{sec:continual_experiments}
To study continual learning, we added a new cargo object to the map, as well as three associated tasks: reach cargo (\textit{Reach}(C)), pickup cargo (\textit{Pickup}(C)), deliver and drop off cargo at the target location (\textit{Deliver}(C)). This experiment consisted of two stages. We first pre-trained our model and two new baselines together on 4 taxi sub-tasks and a new \textit{Reach}(C) sub-task. The \textit{Reach}(C) was included into pre-training to teach the $\oembed$ sub-network to recognitize the cargo object on the map. The learning procedure in this stage was the same as in the multi-task experiment. In the cargo delivery training stage, we added \textit{Pickup}(C) and \textit{Deliver}(C) sub-tasks and fine-tuned the output layers and task embedding $\tembed$ of pre-trained models for $5\times10^6$ steps, keeping all other layers frozen.

To test the utility of the explicit division between task-agnostic episodic memory and recurrent task-dependent policies, we used two baselines:
\begin{enumerate}
    \item \textbf{Baseline (concat)} has the same architecture as Multitask-A2C, but uses the same learning procedure as SEM-A2C (see sec. \ref{sec:meta_rl_procedure}).
    \item \textbf{Baseline (factorized)} is identical to the previous baseline. However, rather than concatenating the task embedding with the LSTM input, we use the same factorization as in $\trnn$ module.
\end{enumerate}

\begin{figure}[htp] 
    \centering
    \includegraphics[width=0.50\textwidth]{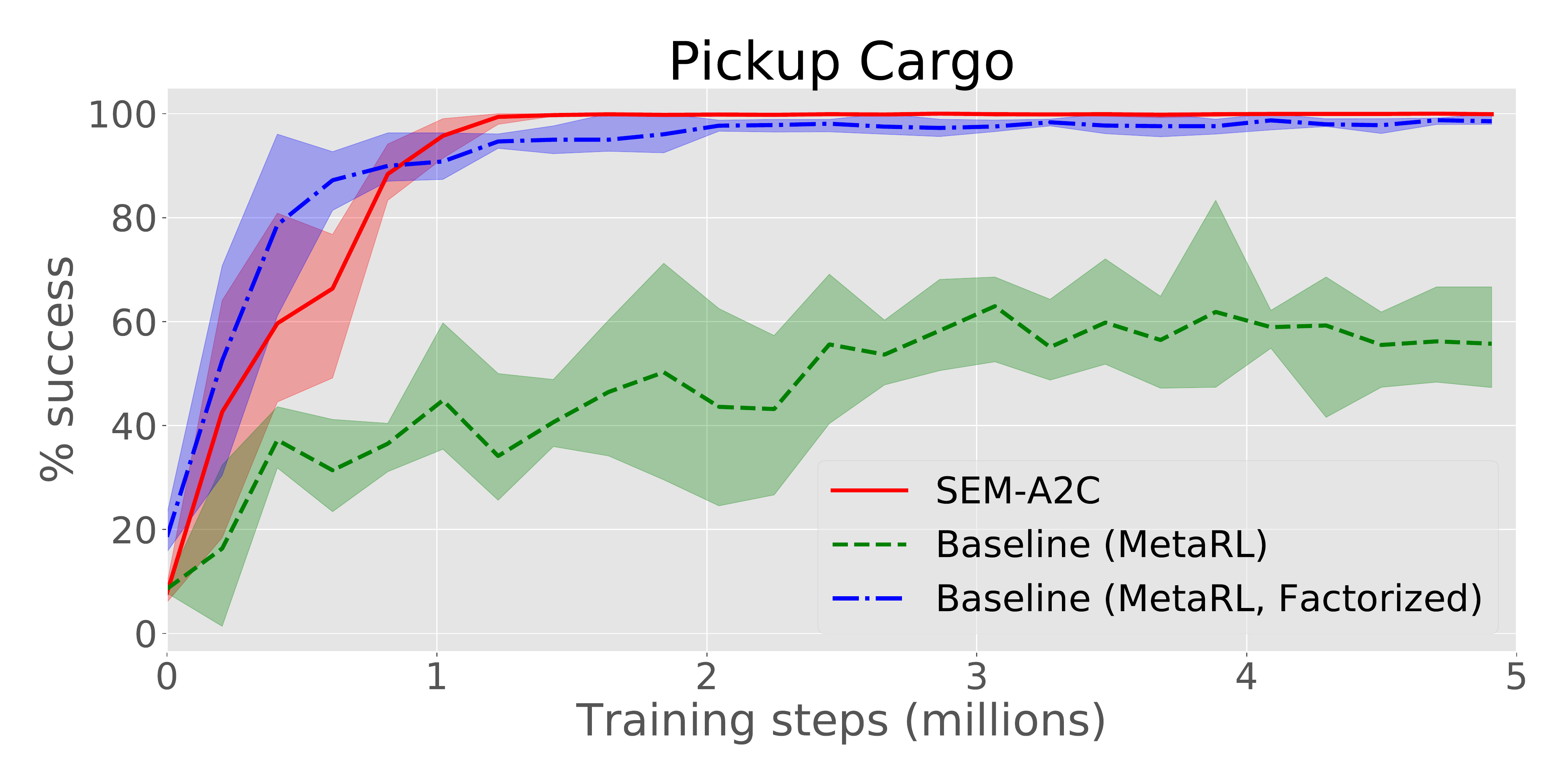}%
    \hfill%
    \includegraphics[width=0.50\textwidth]{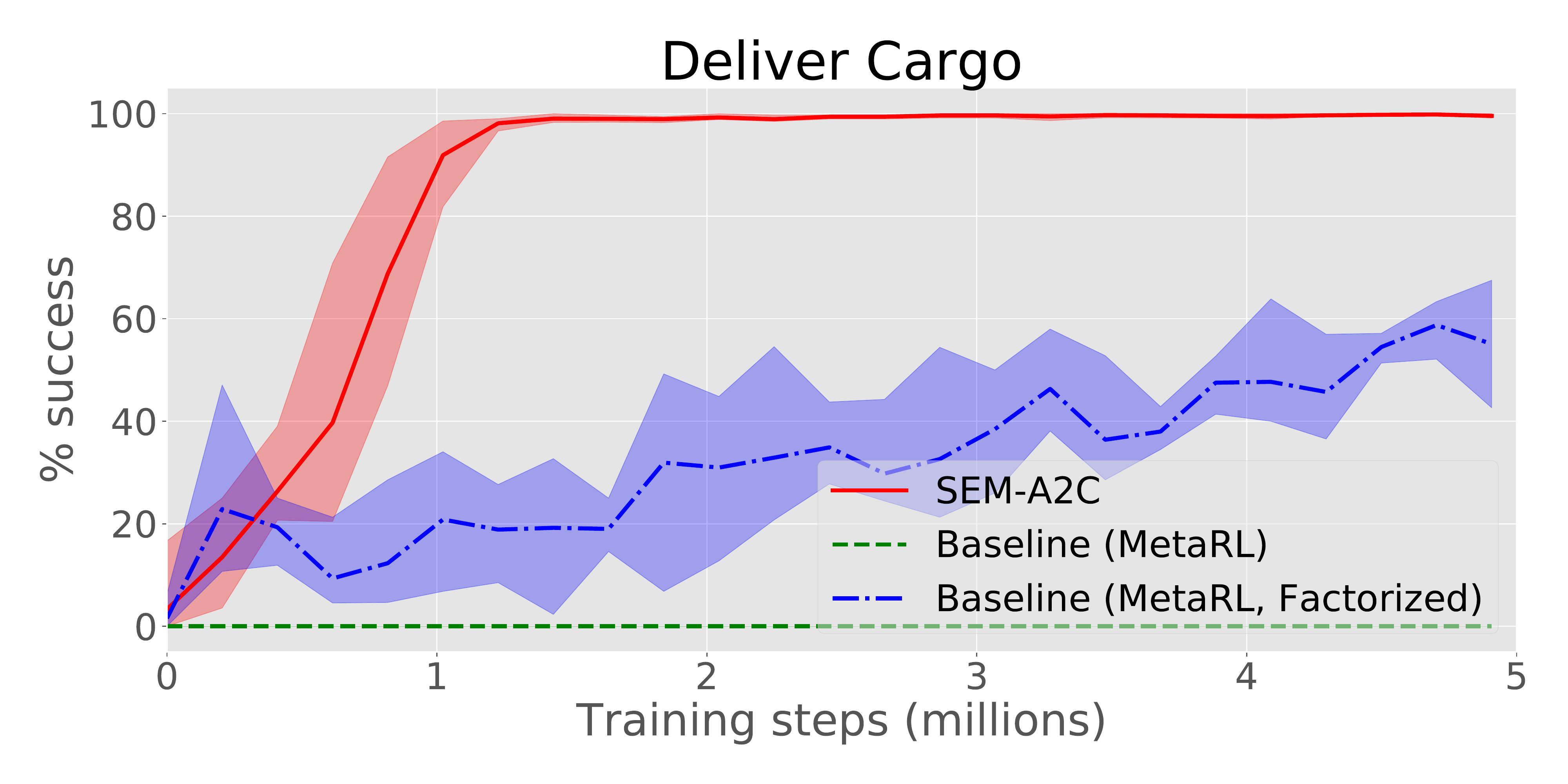}%
    \caption{\textbf{Left}: Success rate for the \textit{Pickup}(C) task during the fine-tuning stage. \textbf{Right}: Success rate for the \textit{Deliver}(C) task during the fine-tuning stage. Each curve is averaged over 6 different runs. The shaded area corresponds to the minimum and maximum scores achieved during these runs.}
    \label{fig:finetuning}
\end{figure}

Figure \ref{fig:finetuning} shows learning curves for new tasks during the fine-tuning stage. Baseline (concat) did not succeed in learning both new tasks. Baseline (factorized) was able to fully learn only the simpler \textit{Pickup}(C) task. On the other hand, SEM-A2C managed to learn both tasks in one million steps.

\section{Conclusions}

Episodic memory helps to solve a significant number of tasks in the real world. Yet in spite of recent progress in the fields of deep reinforcement learning and meta-learning, the question about effective reuse of episodic memory is still open. We proposed and studied a deep neural architecture with shared episodic memory for multi-task problems (SEM-A2C).

The results of our experiments on the Taxi problem demonstrate that our proposed architecture is able to effectively learn how to store and use episodic representation in order to more quickly deliver a passenger. 
SEM-A2C displayed a better performance compared to alternative deep architectures included into the study. We also found that task agnostic episodic memory facilitates acquisition of novel skills for extra tasks in the same environment. Another important result is ability of SEM-A2C to learn sub-task completion. This opens possibility for more autonomous execution of hierarchical tasks by robotic and virtual agents, as a sub-task completion signal can activate the following task in a high level preprogrammed sequence.

We use A2C \citep{a3c_mnih, paac_clemente2017} as a starting point for our modifications and baselines. However, out proposed modifications do not rely on the unique properties of the A2C and can be applied to other general-purpose RL algorithms (PPO, DRQN, etc). 


\subsubsection*{Acknowledgments}

We would like to thank the anonymous reviewers for their suggestions and comments. This work was supported by National Technology Initiative and PAO Sberbank project ID 0000000007417F630002.

\bibliography{drl}
\bibliographystyle{iclr2019_conference}

\appendix

\section{Supplementary materials for "Episodic memory for multi-task and continual reinforcement learning"}

\subsection{Factorized LSTM layer}
\label{sec:f_lstm}
At each time-step, the LSTM layer computes it's output vector $h_t$ and the cell state $c_t$, given the previous vector $h_{t-1}$, the previous cell state $c_{t-1}$, and the current observation $x_t$:
\begin{align}
\begin{pmatrix} i \\ f \\ o \\ g  \end{pmatrix} &= \begin{pmatrix} sigm \\ sigm \\ sigm \\ tanh  \end{pmatrix} T \begin{pmatrix} x_t \\ h_{t-1} \end{pmatrix}, \\
c_t &= f\odot c_{t-1} + i\odot g,\\ 
h_t &= o\odot tanh(c_t),
\end{align}
here $T$ is an affine transformation $T=W*[x_{t}, h_{t-1}] + b$ and $(i,f,o,g)$ are LSTM gates (see \citep{lstm_hochreiter1997}).

For $\trnn$ module LSTM weights $W_{g_t}$ for task $g_t$ are computed as a product of three matrices:
\begin{equation}
W_{g_t} = W_1\operatorname{diag}\left(v_{g_t}\right)W_2, \label{eq:f_lstm}
\end{equation}
here $v_{g_t}$ is an embedding vector of the current task $g_t$. Weights $W_1$ and $W_2$ are shared across all tasks, while task embeddings are trained for each task. This technique allows storing large weight matrices for each task to be avoided. Additionally, the weights factorization significantly increases sensitivity of the $\trnn$ module to different task embeddings. This in turn leads to improved exploration. As shown in Figure \ref{fig:finetuning} the baseline with the factorized LSTM layer outperforms the baseline with regular LSTM layer that gets task embeddings as a part of it's input.

\subsection{SEM-A2C archtecture}
\label{sec:sem_paac_formula}
The following equations describe the forward dynamics of the SEM-A2C network (see fig.\ref{fig:paac_mt_network}):

\begin{align}
\label{eq:o_hat}
\hat{o}_t &= [\oembed(o_t), d_{t-1}, a_{t-1}],\\
\label{eq:h_env}
h^{\env}_t &= \envrnn(h^{\env}_{t-1}, \hat{o}_t),\\
\hat{h}^{\task}_t &= (1 - d_{t-1}) h^{\task}_{t-1},\\
h^{\task}_t &= \trnn(\hat{h}^{\task}_t, \tembed(g_t), [\hat{o}_t, h^{\env}_t]),\\
\pi_t &= \mathit{SoftMax}(\fpol(h^{\env}_t, h^{\task}_t)),\\
v_t &= \fval(h^{\env}_t, h^{\task}_t),\\
\hat{d}_t &= \fcomp(h^{\env}_t, h^{\task}_t).
\end{align}

Training of SEM-A2C is performed in the same way as in A2C \citep{paac_clemente2017} and A3C \citep{a3c_mnih} algorithms. To learn the task completion prediction $\hat{d}_t$ we use cross-entropy loss.

\subsection{Environment Setup}

We studied the proposed model at the modified Taxi problem \citep{max_q_dietterich2000}. The Taxi problem is a clear example of a task with a hierarchy of sub-tasks. The goal of this problem is to maneuver a taxi in order to reach a passenger placed in a random location, and then to pick up and deliver the passenger to a target location. Thus the main task of delivering a passenger from an initial to a target location, is divided into 4 separate sub-tasks:

\begin{enumerate}
\item Reach (P): to get to the passenger on the map;
\item Pickup (P): to put the passenger in the car;
\item Reach (D): to transport the passenger to the target location;
\item Dropoff (P): to disembark the passenger.
\end{enumerate}

In the original Taxi game, a map on which the agent operates was fixed throughout the training. In our implementation, the map is created randomly for each new episode. On the map, $10\%$ of the blocks are filled with impassable terrain (walls) and another $10\%$ contain difficult terrain (water). The starting positions of the agent, the passenger and the target are chosen randomly. In order to increase the difficulty of the problem, the agent is only permitted to view the small 7 by 7 area of cells surrounding it. The agent does not possess knowledge of the initial coordinates of the passenger and the target. It must independently find them on the map. For each action, the agent receives a penalty of $-0.1$, and any action performed on difficult terrain the penalty increases to $-0.3$. After sub-task completion, the agent receives a reward equal to $1$. The length of one training episode is 400 steps. Our implementation is built on top of the MazeBase engine \citep{mazebase_sukhbaatar2015}.

\end{document}